\documentclass{article}

\usepackage{ijcai18}
\usepackage{times}
\usepackage{helvet}
\usepackage{graphicx}
\usepackage{url}

\usepackage{amsmath}
\usepackage{amssymb}
\usepackage{amsthm}
\usepackage{enumerate}

 \usepackage[small]{caption}

\usepackage{xspace}
 \usepackage{ifthen}
 \usepackage{tikz}

\newcommand{\Omit}[1]{}

\newcommand{\citeay}[1]{\citeauthor{#1}\,[\citeyear{#1}]}

\newcommand{\bi}{\begin{itemize}}
\newcommand{\ei}{\end{itemize}}
\newcommand{\be}{\begin{enumerate}}
\newcommand{\ee}{\end{enumerate}}

\begin{document}

\title{Model-free, Model-based, and General Intelligence}

\author{
Hector Geffner$^{1,2}$\\
$^1$ Universitat Pompeu Fabra, Roc Boronat 138, 08032 Barcelona, Spain \\
$^2$ ICREA, Pg. Llu\'{\i}s Companys 23, 08010 Barcelona, Spain \\
hector.geffner@upf.edu
}


\pdfinfo{
  /Title (Model-free, Model-based, and General Intelligence)
  /Author (Hector Geffner)
  /Keywords (planning, learning, general intelligence)
  /Subject (artificial intelligence)
}

\maketitle

\begin{abstract}
During the 60s and 70s, AI researchers explored intuitions about intelligence by 
writing programs that displayed  intelligent behavior. Many good ideas came out
from this work  but  programs written by hand were not robust or general.
After the 80s, research  increasingly shifted  to the development of \emph{learners}
capable of inferring  behavior and  functions from experience and data,
and \emph{solvers} capable of tackling well-defined but intractable models like SAT,
classical planning, Bayesian networks, and POMDPs. The learning approach has
achieved considerable success but results in black boxes that do not have the
flexibility, transparency, and generality of their model-based counterparts.
Model-based approaches, on the other hand, require models and scalable algorithms.
Model-free learners  and  model-based  solvers  have close parallels
with   \emph{Systems~1 and 2} in current theories of the human mind: the  first, a fast, opaque,
and inflexible intuitive mind; the second,  a slow, transparent, and flexible analytical mind.
In this paper, I review developments in AI and draw on these theories 
to  discuss  the gap between model-free learners and model-based solvers, 
a gap that  needs to be bridged in order to have intelligent systems
that are robust and general.
\end{abstract}

\section{Introduction}

The current  excitement about AI  is the result of a number of breakthroughs in machine learning.
Some of these developments are true AI milestones,  like the  programs that
achieve top world performance in games such  as Chess and  Go by
learning from  self-play only  \cite{silver2,silver2017mastering}. 
These are no small accomplishments and most of them  have to do with 
deep learning \cite{lecun2015deep} and deep reinforcement learning \cite{dqn}.
The goal of this paper is to place these developments in perspective,
in particular by comparing  \emph{model-free learners} with \emph{model-based solvers}.
Solvers are programs that accept instances of different types of models, 
like SAT, classical planning, Bayesian networks, and POMDPs, and automatically
compute their solutions.  Learners and solvers have interesting similarities
and differences. Both   are concerned with the derivation of functions $f$
for mapping inputs $x$ into outputs, but while learners derive the function  $f$
from data or experience, solvers compute the output $f(x)$ for each given input $x$
from a model.  Interestingly the scope and characteristics of  model-free learners and
model-based solvers have  close parallels with those of the two systems or processes
that are currently assumed to  make up the human mind  \cite{fast-slow,dual-mind}: one, a  fast, opaque, and
inflexible intuitive mind (System 1); the other, a slow,  transparent,  and  general analytical mind (System 2).
A number of works have analyzed the limitations of deep learning,  often placing the emphasis on the need to construct and
use \emph{models} \cite{josh,darwiche,pearl,marcus1}. In this paper, I place the emphasis  on \emph{solvers}, and in particular \emph{planners}, 
which require both models and general, scalable algorithms. The Systems 1 and 2 perspective will be  useful for understanding the scope of
learners and solvers, and the need for a tight integration.  

The paper is organized as follows. I trace some history, look at learners,  solvers, and planners, which are a particular type of solvers, 
and  address  the similarities and differences between learners and solvers, and  the challenge of integrating them. 

\section{Programming, AI, and AI Programming}

Artificial Intelligence is a brain child of Alan Turing and his universal computer \cite{turing}.
Turing was not only a logician and the father of the modern computer but also the first
modern programmer. Programming played  a key role   in the early days of AI
and in the selection of papers for  the seminal  \emph{Computers and Thought} book
where each chapter  described  ``a working computer program'' \cite{computers-and-thought}.
The  emphasis on working programs, however,  was not shared by everyone.
John McCarthy,  who was both the father of the most popular AI programming language at the time  (Lisp)
and the person  that  gave AI its name, was  putting the basis for the logical approach to AI
where the emphasis was  not in the use of representations but in their  expressiveness and semantics
\cite{mccarthy:logic}. 
Yet such dissenting views were not common and are not to be found in the CT book. 
Several of the key  AI contributions  in 60s, 70s,  and  80s  had to do indeed  with \emph{programming}
and the \emph{representation of knowledge in programs}, and this includes   Lisp  and functional programming,
Prolog and logic programming, rule-based programming, expert systems  shells, frames, scripts, and semantic networks
\cite{ai-programming:mcdermott,ai-programming:norvig}.

\section{The Problem of Generality}

Many great ideas came out of this  work but  there was a  problem:
the programs were not sufficiently robust or general,  and they tended to fail
on examples or scenarios not anticipated by the programmer. It was natural 
to  put the blame  on the knowledge that the programs were missing but it was not
clear then   how to proceed.  Some programmers  narrowed down the domain and scope of the programs
so that all the   relevant knowledge could be made  explicit. This was the 'expert systems' approach.
Some  took the programs as illustrations or demonstrations of potential capabilities
which  were not  actually delivered.  Some decided to sit down and  write down  all the relevant commonsense knowledge.
This was the motivation underlying projects like  CYC \cite{cyc}. McCarthy  discussed related
issues in his 1971 ACM Turing lecture entitled ``Generality in Artificial Intelligence'',
published in revised form many years later \cite{mccarthy:generality}.

The limitations of AI programs for exhibiting  general intelligence not tied to toy worlds or narrow domains
led to an impasse  in  the 80s, one of whose effects was  a \emph{methodological shift}  in which  research increasingly
moved away from  \emph{writing programs for ill-defined problems}  to    \emph{designing algorithms for well-defined  mathematical
tasks}. The  algorithms  are general in the sense that they are not tied to particular examples but to certain classes of models and tasks
expressed in mathematical form. The new programs, \emph{learners} and \emph{solvers},  have  a crisp functionality and both can  be  seen as computing
\emph{functions} that map  inputs into outputs (Figure~\ref{fig:solvers}).

\section{Learners}

The current excitement about AI in academia and in industry  is the result of a number of breakthroughs 
that have pushed  the state-of-the-art in  tasks such as  image understanding, speech recognition, and challenging games
\cite{krizhevsky2012imagenet,graves2013speech,dqn}, in certain cases delivering superhuman performance 
while starting with basic knowledge of the rules of the game only  \cite{silver2017mastering}. These developments have to do with two classes of
learners: \emph{deep learners} and \emph{deep reinforcement learners.} While these are not the only approaches pursued in machine learning,
they are the ones behind these milestones and the ones on which I focus.

In both deep learning (DL) and deep reinforcement learning (DRL), training results in a function $f_{\theta}$
that has a fixed structure, given by a  deep neural network \cite{lecun2015deep}, and  a number  of  adjustable parameters $\theta$.
In DL, the input vector $x$ may represent an image   and the output  $f_{\theta}(x)$, a classification label, a probability distribution over
the  possible labels, or object labels with suitable bounding boxes. In DRL,  the input $x$  may represent the state of a
dynamic system or game, and the output $f_{\theta}(x)$,  the value of the state, the action to be done in the state, or a probability distribution over such actions.
The key difference between DL and DRL is in the way in  which the functions $f_{\theta}$ are learned during training. Deep learning is a supervised method where the
parameters $\theta$ are learned by  minimizing  an error  function that  depends on the  inputs and target outputs
in a training set. Deep reinforcement learning, on the other hand,  is a non-supervised method that learns from experience,
where the error function  depends on the value of states and their successors. In both DL and DRL, the  most common
algorithm for minimizing the error function is stochastic gradient descent   where the parameter vector $\theta$ is modified
incrementally by taking  steps  in the direction of the gradient.  Similar optimization  algorithms are used in policy-based DRL
where the function $f_{\theta}$ represents a policy.

The basic ideas underlying DL and DRL  methods are not new and can be traced back to the neural network and reinforcement learning
algorithms of the  80s and 90s \cite{pdp,lecun:1989,sutton:book}.
The recent successes have to do with  the gains in  computational power and  the ability to use deeper nets on more  data.  The use of these
methods on problems that have attracted and spurred  commercial interest has  helped as well. A common question that arises  is what are
the limits of these methods. One important restriction is  that \emph{the inputs $x$  and outputs $y$ of neural nets have a bounded, fixed size.}
This limitation  is not relevant for learning to play chess or Go whose boards have a fixed size, but is relevant for tackling, for example,
arbitrary instances of the much simpler Blocks world. Attention mechanisms have been proposed as a way to deal with arbitrarily large inputs
\cite{attention}, but such mechanisms  introduce partial observability, which is another challenge in learning.  Indeed,  as  one  moves  away from the
basic setting of DL and DRL, the  methods,  the results, and the evaluation standards, all  look somehow weaker. 

\begin{figure}[t]
  \begin{center} \textit{Input}  $x$  $\Longrightarrow$ \fbox{\textsc{Function} $f$}   $\Longrightarrow$ \textit{Output}  $f(x)$ \end{center}
  \caption{\small Learners and solvers map inputs into outputs. Learners derive  the function $f$  from data or experience.
    \emph{Solvers} derive  the value of the  function $f(x)$ for each given input $x$ from a model.}
\label{fig:solvers}    
\end{figure}

\section{Solvers}

The second type of programs that we consider are \emph{solvers}.   Solvers  take  a convenient description of
a particular model instance (a classical planning  problem, a constraint satisfaction problem, and so on) and automatically compute its solution.
Solvers can also be thought as computing a function $f$ mapping inputs $x$ into outputs,  the difference is that
they  work out of the box  without training  by computing the output $f(x)$ lazily for each  given input $x$. 
The  target  $f(x)$ is given implicitly by the  model. For a SAT solver, the inputs  $x$ are 
formulas   in conjunctive normal form, and the output $f(x)$ tells whether the formula $x$ is satisfiable or not. 
For a classical planner, the inputs are classical planning problems $x$ and the output is a plan that
solves the problem.  Solvers have been developed for a variety of models that  include constraint satisfaction problems (CSPs),
SAT, Bayesian networks, classical planning, fully and partially observable non-deterministic and stochastic planning problems
(FONDPs, PONDPs, MDPs, and POMDPs), general game playing, and answer set programming among others
\cite{dechter:book,sat:handbook,pearl:book,geffner:book,bertsekas:dp,cassandra:pomdps,ggp,asp}.
Solvers are \emph{general} as they must deal with \emph{any} problem that fits the model: any classical  planning problem, any
CSP, etc. This presents a crisp  \emph{computational challenge}, as all  models  are computationally intractable, 
with complete algorithms running in time that is exponential in the number of problem variables or worse. 
The  challenge is to push this  exponential explosion  as far as possible, as 
solvers should not break down on a problem merely because it  has many variables. 
To achieve this,   domain-independent solvers must be able to   exploit the structure of the given problems so that their
performance over a given domain can approach the performance of a domain-specific solver.
The computational value of solving techniques is assessed experimentally and  in most   cases  by means of competitions. Competitions
have helped to generate hundreds of problems  used as benchmarks, have set standards for the  encodings  of problems,  and
have facilitated  the empirical evaluation of algorithms.  The focus on models and solvers that can scale up has acted
as \emph{a powerful filter on ideas and techniques}, setting up a clear distinction between the ideas that look well from
those that actually work well. We illustrate some of these ideas in the context of planning to  make the comparison
between learners and solvers more concrete. 


\section{Planners} 

Planners are solvers for models that  involve  goal-directed behavior.
Planners and planning models come in many  forms depending on a number of dimensions including:
1)~uncertainty about the initial situation and action transitions, 2)~type of sensing, 3)~representation of uncertainty,
and 4)~objectives. The simplest type of planning, classical planning, involves no uncertainty
about the initial situation or action effects, hence no uncertainty or sensing, and the objective
is to reach some condition. MDPs have stochastic actions and full state observability
and the   objectives are  described in two different ways: as goals to be reached with probability
1 by applying actions with positive costs (goal MDPs), or as discounted rewards to be
collected (discounted reward MDPs).
Other types of objectives considered in planning are 
temporally extended goals,  like ``visit rooms 5 and 7 forever'', that are conditions on
possibly infinite state trajectories  \cite{camacho:ltl}. 
Three  other relevant dimensions in planning are whether 5)~models
are expressed  in compact or flat form, 6)~solutions are sought
off-line or on-line, and in the latter case, 7)~which off-line solution form is sought. 
About the last point, partially observable problems can be solved optimally
with policies that map belief states into actions but can also be approached effectively
with simpler  solution forms such as   finite-state controllers.
Many of these dimensions affect the complexity of planning.
Focusing on the decision problems, classical planning  is PSPACE-complete,
while classical planning with a fixed horizon is NP-complete \cite{kautz:satplan}.
Similarly,   PONDPs expressed in compact form  are  2-EXP complete \cite{rintanen:po},
while computing  memoryless policies for PONDPs expressed in flat form  is  NP-complete  \cite{chatterjee:SAT}.

\Omit{
Three  clarifications before proceeding. Planning is related to but is different than  model-based RL, 
where some models parameters  are unknown. Similarly, Monte-Carlo Tree Search (MCTS) \cite{uct} is a popular
planning algorithm but not  the only or best planning algorithm in general. Finally, hierarchical task network (HTN)
planners
are not solvers in our sense, as they are  used to encode strategies for solving problems
rather than problem descriptions. There are however model-based RL methods that reduce RL to planning \cite{r-max},
MCTS  can be understood as a variant of the  traditional AO* algorithm  \cite{bonet:uct},
and the automatic derivation and exploitation of  hierarchies  in planning
is an important but still open problem.
}


We look next at three powerful computational  ideas  in planning, relaxations, transformations, and width,
drawing in each case contrasts  between planners and  learners.

\subsection{Relaxations}

The derivation of heuristics in classical  planning for guiding the search for plans 
follows an  old idea: if you do not know how to solve a problem, solve a simpler problem 
and use its solution as guidance for  solving the original problem \cite{pearl:heuristics}.
In planning, the most useful simplification $P'$ of a planning problem $P$
is  the \emph{monotonic relaxation}  where the effects of the actions on the variables are made monotonic  \cite{mcdermott:h,bonet:h}.
For this, the states $\bar{s}$ of the relaxation $P'$  are defined not as assignment of values $x$ to variables $X$ but as
collection of atoms $X\!=\!x$. In the original problem  $P$, if an action $a$  makes an atom $X\!=\!x'$ true, it
makes the  atoms   $X\!=\!x'$ for $x\not\!=\! x'$ false. In the monotonic relaxation, on the other hand,
the new atom $X\!=\!x'$ does not make the other atoms false. A result of this is that the monotonic relaxation  $P'$ becomes
\emph{decomposable}: a plan that achieves two atoms $X\!=\!x$ and $Y\!=\!y$ jointly in $P'$ can be obtained
by concatenating  a plan   for $X\!=\!x$ and a plan for $Y\!=\!y$. This idea can be used to  compute plans  for all atoms and hence for
the conjunction of  atoms  in the goal  of  $P'$  in low polynomial time.  The heuristic $h(s)$ can  be set then  to the
size of the plan in the relaxation obtained from the state $\bar{s}$ where the only true atoms are those which are true in $s$ \cite{hoffmann:ff}.\footnote{
When variables are boolean and actions are represented in STRIPS as precondition, add, and delete lists,
the monotonic relaxation is known as the \emph{delete-relaxation}, as the monotonic relaxation is obtained by making all delete
lists empty.}
Planners compute the  heuristic value $h(s)$ for any state $s$  \emph{over any problem} $P$  following variations of this
procedure. This is why they are called \emph{domain-independent} planners.  Learners can infer the  heuristic function $h$ over all the states $s$ of
a \emph{single} problem $P$ in a straightforward way, but they can't  infer an heuristic function $h$ that is  valid for \emph{all problems}.
This is natural: learners can deal with new problems only if they have acquired experience on related problems. 

\Omit{
  Current planners  use the delete-relaxation  to compute heuristic,  to derive  implicit intermediate goals, called landmarks \cite{hoffmann:landmarks,richter:lama}
and to identify the actions that appear most relevant in the state $s$, called the helpful actions \cite{hoffmann:ff}.
}

\subsection{Transformations}

Planners tend to work well even  in large  classical instances yet  many  problems do not have this form.
Transformations are  used to leverage  on  classical planners. 

\medskip

\begin{figure}[t]
\begin{center}
  \begin{tabular}{ccc}
    \includegraphics[width=1.3in]{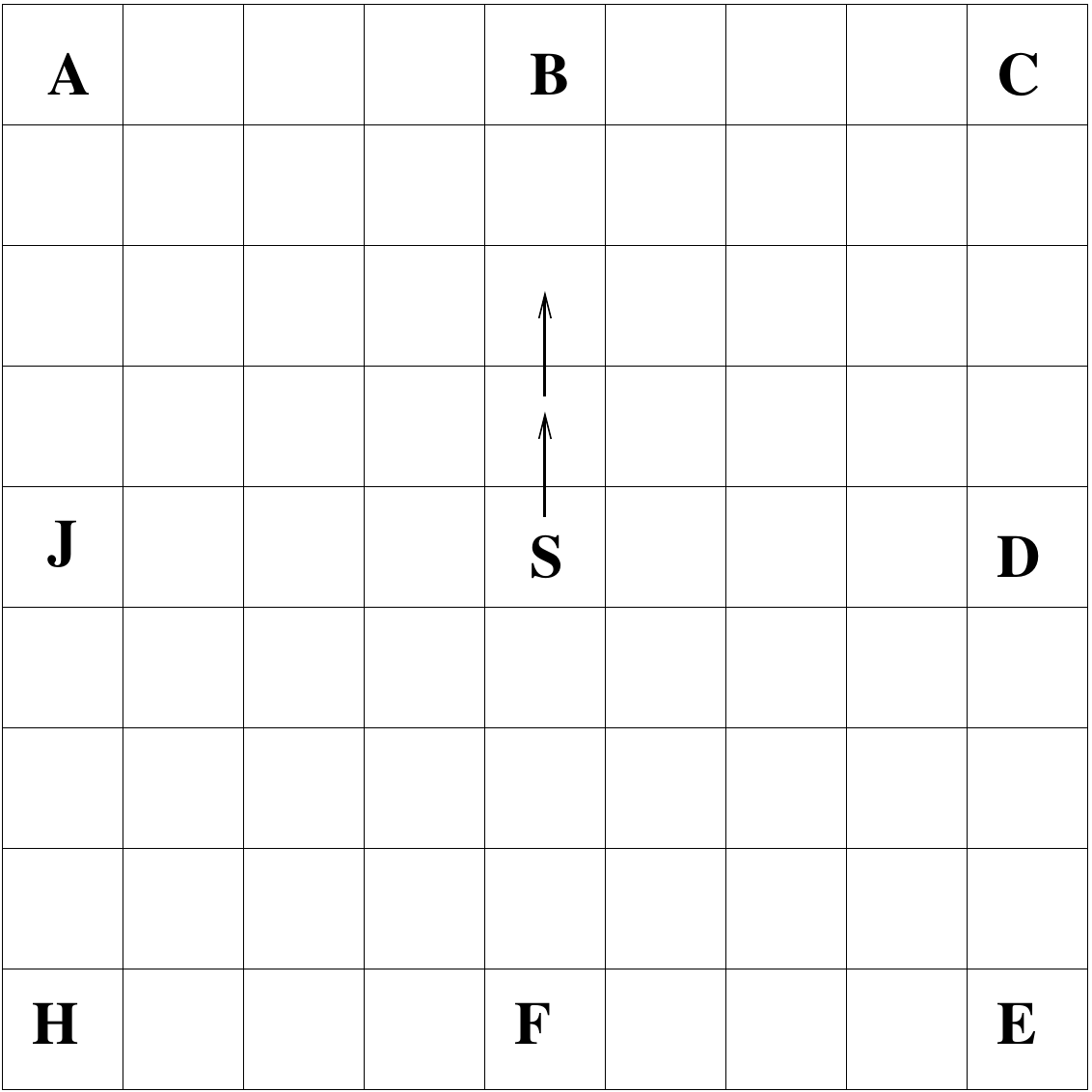} & \hspace{0.2cm} & 
        \includegraphics[width=1.3in]{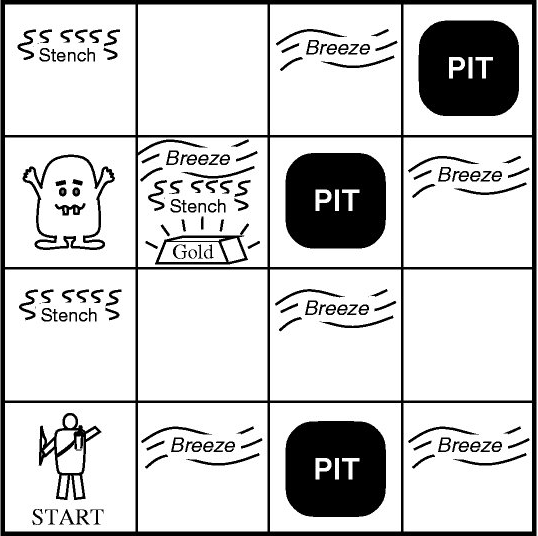}
    \end{tabular}
\end{center}
\vskip -0.15cm
\caption{\small \emph{Left.} Goal recognition: where is the agent heading to?
  \emph{Right.} Partial  observability: get the gold without knowing
  where the gold, pits, and wumpus are, using the sensors (see text).}
\label{fig2}
\end{figure}

\noindent \textbf{Goal recognition.} Goal recognition is a \emph{classification problem}
where the hidden  goal of an agent has to be uncovered from  his observed behavior and a pool
of possible goals. For example,  an agent is observed to  move  up twice from the middle of the grid
in Figure~\ref{fig2}, and we want to determine to which of the marked targets he may be heading to.
A  goal recognition problem $P({\cal G},O)$ can be represented as a classical   problem $P$ but 
with the goal replaced by a pool $\cal G$ of possible goals $G$ and a sequence of observed actions $O$.
In the example, the reason  that the observed agent is likely to be moving  to one of  targets in the top row is that
it is \emph{not} reasonable (cost-efficient)  to reach the other targets by moving up.
This can be formalized  in terms of \emph{two cost measures} (plan lengths) for each possible goal $G$:
the cost of  reaching $G$ with a plan that complies with the observations, denoted $c^+(G)$,
and  the cost of reaching $G$ with a plan that does not  comply  with the observations, denoted $c^-(G)$.
Bayes' rule says that the goal posterior is $P(G|O)=P(O|G) \, P(G)/P(O)$ where the priors $P(G)$ can be assumed to be given. 
The posterior $P(G|O)$ is then  determined  by the likelihoods  $P(O|G)$ that express how well each of the possible goals
$G$ predicts  $O$. In \cite{ramirez:aaai2010}, the likelihood $P(O|G)$  is set to a monotonic function $f$ of the cost difference
$\Delta(G) = c^-(G) - c^+(G)$ that expresses  that the costlier that it  is to  reach $G$ with a plan that complies with $O$
in  relation to one that does not, the less likely that it is the observation $O$ given $G$. 
The  costs are computed  by running a  planner on  problems obtained  from  the goal recognition problem $P({\cal G},O)$.
The posterior distribution $P(G|O)$ for all goals  $G \in {\cal G}$  is  computed using Bayes' rule by calling a  classical planner
$2 |{\cal G}|$ times. Once again, even if  \emph{goal recognition is a classification problem}, it is not possible to obtain  a  \emph{general, domain-independent}
account of goal recognition using learners instead  of planners. 


\medskip

\noindent \textbf{Partial observability.}  Figure~\ref{fig2} shows an instance of the Wumpus problem \cite{russell:book}  where an agent
has to get the gold while avoiding the monster and the pits. The positions of the gold, the monster, and the pits, however, are not known to the
agent  that can sense  stench if at distance one from the wumpus, a breeze, if at distance one from a pit, and a shiny object if in the same cell
as the gold. The problem can be modeled as a POMDP or as a  contingent planning problem but few  planners  scale up when the size of the grid is
increased. Even if we focus on \emph{on-line planning} methods, two challenges need to be addressed: tracking the beliefs about the true hidden state,
and selecting the  actions that take the agent to the goal or provide relevant information. 
The two challenges have been addressed jointly through a  ``planning under optimism'' scheme
where  the partially observable problem $P$  is relaxed into a classical problem
$K(P)$ inside a \emph{plan-execute-observe-and-replan} loop \cite{bonet:aaai2014}. The transformation $K(P)$ removes 
the uncertainty in $P$  in two steps. First,  literals $L$ that represent that $L$ is true are replaced by two literals $KL$ and $K\neg L$,
the first meaning  $L$ is known to be true, the second meaning  that $L$ is known to be false.
Second, sensing the truth value of a positive literal $L$ that is currently unknown is mapped into
two actions:  one that  predicts  $KL$, and another that predicts $K\neg L$.
The \emph{plan-execute-observe-and-replan} loop then executes the prefix of the  plan obtained from $K(P)$
using a classical planner until a discrepancy is found between what has been predicted and 
what has been observed.  In such a case, the observed literals are added, a new plan
is computed, and the loop continues until the goal is reached. For problems with no dead ends and \emph{belief width} 1,
the executions are guaranteed to reach the goal in a  bounded number of steps \cite{bonet:aaai2014}.
Problems like  Wumpus over 15x15 grids  are solved consistently very fast.
On the other hand,  coverage over instances of problems like Minesweeper is not 100\%, which is natural
as belief tracking in  Minesweeper is NP-hard \cite{minesweeper:np} while incomplete and  polynomial in the planner
\cite{palacios:jair}. Learning approaches face two challenges in these types of problems: dealing with partial observability
over instances of the same domain, and more critically, transferring useful knowledge from instances of one domain
to instances of another.

\Omit{
(tracking beliefs with the $KL$ literals is sound but incomplete  \cite{palacios:jair}).  Methods  that leverage on classical planners for
planning with MDPs,  FONDPs, and partial observability are reported in \cite{ff-replan,muise:prp,shani:po}.
}

\medskip

\begin{figure}[t]
\centering
\medskip
\includegraphics[width=3in]{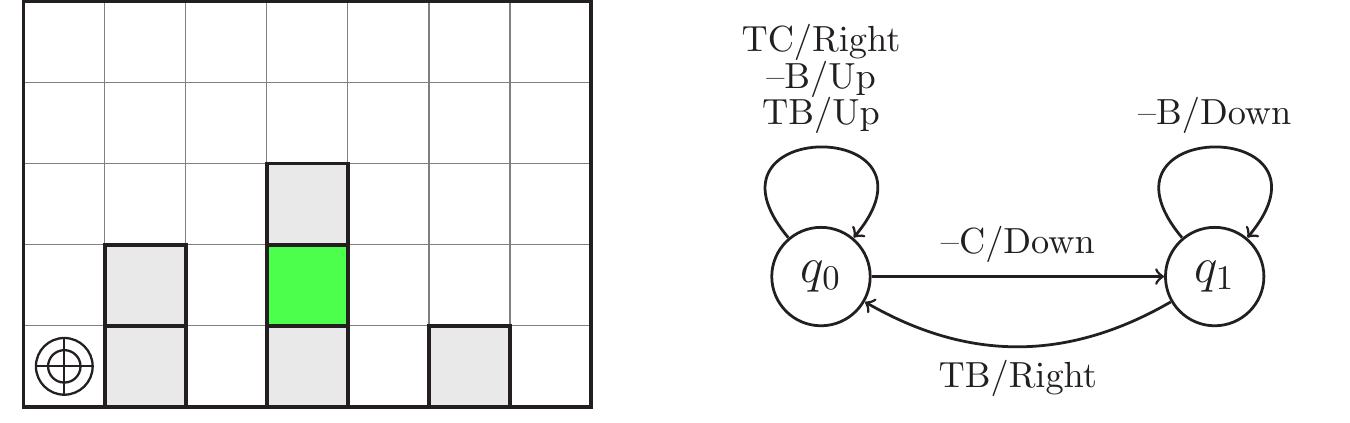}
\caption{\small \textit{Left:} Problem where a visual-marker (mark on the lower
  left cell) must be placed on top of a green block whose location is not known, 
  by moving the mark one cell at a time, and by observing what's in the marked cell.
  \textit{Right:} Two-state controller obtained with a classical planner. 
  The controller solves the problem and any variation resulting from changes in the number or configuration of blocks.
  Edge $q \rightarrow q'$ labeled $o/a$ represents tuple $(q,o,a,q')$. 
}
\label{fig:beyond:visual-marker}
\end{figure}

\noindent \textbf{Finite-state controllers and generalized planning.} Finite-state controllers represent an  action selection
mechanism widely used in video-games and robotics. A finite-state controller for a partially observable problem $P$ with
actions $a$ and observations $o$ can be  characterized by a set of tuples $(q,o,a,q')$ with no pair of tuples sharing
the first two components  $q$ and $o$.  Each such tuple  prescribes the  action $a$ to be done  when the controller state is $q$
and the observation is $o$,  switching  then to the controller state $q'$ (which may be equal to $q$ or not).
A  controller solves  $P$ if starting in the distinguished controller state $q_0$, all the executions
that are compatible with the controller reach a goal state. In the approach by \citeay{bonet:icaps2009}, the problem $P$ is transformed
into a classical planning  problem $P'$ whose actions are associated with each one of the possible tuples $(q,o,a,q')$,
and where extra fluents $p_q$ and $p_o$   track the controller states and observations. The action $t=(q,o,a,q')$ behaves
like the  action $a$ but conditional on  $p_q$ and $p_o$ being true, and  setting $p_{q'}$ to true. 
Figure~\ref{fig:beyond:visual-marker}  shows a problem where a visual-marker or eye (circle on the lower left) must be placed on top of a green block by
moving it one cell at a time. The location of the green block is not known, and the observations are whether the
cell currently marked contains a green block (G), a non-green block (B), or neither (C), and whether this cell is at the level
of the table (T) or not (--). The finite-state controller shown on the right has been computed by running a classical planner over the transformed problem. 
The controller moves the eye from left to right at the level of the table, and when it finds a block that is not green,
it moves the eye up the tower and then down.  Interestingly, this controller not only solves the problem shown on the left,
but  \emph{any modification of  the problem resulting from changes in the number or configuration of blocks}. This is an instance of
what has been called \emph{one-shot learning}: the controller obtained by the classical planner for one instance  solves  all instances of the problem.
The task  of computing solutions to families of planning problems, for example, all  instances of Blocks world featuring  blocks $A$ and $B$ where the goal 
is $A$ on $B$,  is called \emph{generalized planning} \cite{srivastava:aij2011,hu:generalized,bonet:ijcai2017}. Generalized  planning provides an interesting
bridge  to  work in DRL \cite{siddartha:sokoban} as a generalized plan is not just the output $f(x)$ for a  particular planning instance $x$
(Figure~\ref{fig:solvers}), but the function $f$ itself. This is indeed a case where \emph{planners and learners
aim at the same type of plans}, ones using models, the others, experience. 

\subsection{Width}

The third and last idea that we consider in planning is related to
a very simple search algorithm called IW($k$) that has some remarkable
properties \cite{nir:ecai2012}. The algorithm assumes that the states $s$ assign values to a number of boolean features $F$
that are given. In classical planning, the boolean features are the atoms of the problem
but the algorithm applies to other settings as well. For $k=1$, 
IW(1)  is a standard \emph{breadth-first search}  with just one change:
if upon generation of a state $s$, there is no feature $f_i$ in $F$ such that $f_i$ is true in
$s$ and false in all the states generated before $s$, the state $s$ is pruned.
In other words, the only states that are not pruned are those that make \emph{some} feature  true for the first time.
Such states are said to have novelty $1$. The number of states expanded by  IW(1)  is thus  \emph{linear} in $|F|$
and not exponential in $|F|$ as in  breadth-first search. The algorithm IW($k$) is  IW(1) but applied to
the larger feature set $F^k$ made up of the conjunctions of  $k$ features from $F$.
A basic property  of IW($k$) is that most  classical planning benchmark domains
can be formally shown to have  a bounded and small \emph{width} $w$ no greater than 2 when goals are \emph{single atoms},
meaning that any such     instances  can be  solved optimally (shortest plans) by running IW($w$) in low polynomial time.
For example, the goal $on(x,y)$ for any two blocks $x$ and $y$ in Blocks world
can be shown to have width no  greater than $2$, no matter the number of blocks or  initial configuration. This means that IW(2) finds a
shortest plan to achieve $on(x,y)$ in polynomial time in the number of atoms and blocks even though the state space for the problem
is exponential. The majority of the benchmarks, however, do not feature a single atomic goal but a conjunction of them, and a number of
extensions of IW($k$) have been developed aimed at them, some of which represent the current state of the art
\cite{nir:aaai2017}. Moreover, unlike all the classical planning algorithms developed so far,  including those that rely on
heuristics obtained from relaxations or  reductions into SAT, some of these extensions scale up equally well
without taking into account the  structure of actions (preconditions and effects), meaning that they 
can plan  with \emph{simulators}  \cite{frances:ijcai2017}. In particular, IW(1) has been used effectively
as an on-line planner in the Atari video games in  two modes: with  the memory states of the emulator \cite{nir:ijcai2015}
and with the   screen states \cite{bandres:aaai2018}.  In the first case, the  features $f_i$ are  associated
with each of the 256 values of  the 128 memory bytes; in the second case, the features are the B-PROST  pixel
features \cite{shallow} defined and motivated by the neural net architecture  underlying   DQN \cite{dqn}.

\section{Learners and Solvers: Contrasts}

The IW(1) planner and the DQN  learner perform comparably  well in the Atari games
while working in ways that illustrate \emph{key differences between  learners and  solvers:}
DQN requires lots of training data and time,  and then plays very fast, reactively; IW(1)  plays out of the box   with no  training
but needs to think a bit before each move.\footnote{0.5 seconds per move in the rollout version of IW(1).}
This is a   general characteristic: learners require training, which is often slow, but then are fast;
solvers can deal with new problems with no training   but after  some deliberation.
Solvers are thus general, domain independent as they are called, in a way that learners can't be:
solvers can deal with new problems from scratch, provided a suitable representation of the problems; 
learners  need experience on related problems. 

The differences between \emph{model-free learners} and \emph{model-based solvers} are reminiscent of  current accounts
in psychology that describe the human mind as made of two interacting systems or processes:
a System 1 associated with the  ``intuitive  mind'', and a  System 2 associated with the  ``analytical  mind'' \cite{dual-mind}.
\citeay{fast-slow} refers to the two processes  as fast and slow thinking.  Common characteristics associated 
with these systems are:

\begin{center}
  \begin{tabular}{ccc}
    \textbf{System 1} & \hspace{.cm} & \textbf{System 2} \\[.1cm]
    fast     &&       slow \\
    associative &&   deliberative \\
unconscious   && conscious \\
effortless   &&  effortful \\
parallel &&       serial \\
automatic  &&     controlled \\
heuristic   &&    systematic \\
specialized  &&    general \\
\ldots & & \ldots 
  \end{tabular}
\end{center}

It is hard not to see the parallels between the characteristics associated with Systems 1 and 2 on the one hand,
and those of  learners and solvers on the other. The processes underlying Systems 1 and 2 in the human mind, however,
are not independent: System 2 is assumed to have evolved more recently than System 1 and to be based  on it.
For  activities like solving a Sudoku, writing a proof, or understanding a long text, it is assumed that the analytical System 2
is in command, integrating the suggestions of System~1 that are  triggered by clues issued by System 2 or  picked up from context.
Our account of classical planning as heuristic search has this flavor: the deliberate search for the goal is informed by  an
heuristic  which must be  opaque to the ``analytical mind'', as it is computed from a relaxed model that has  no resemblance to 
the real world  \cite{geffner:pearl-heuristics}.
More generally, there cannot be a  System 2 without a System 1,  as reasoning is computationally hard and for inference methods to be effective
they have to be ``biased'' to exploit the structure of  common tasks. At the same time, human cognition cannot be a System 1 process only, 
as there are  situations and new problems where accumulated experience and built-in mechanisms are not  adequate.

\section{Learners and Solvers: Challenges}

The learning algorithm AlphaZero at the heart of the systems that  learn  to
play chess and Go at  world-class level by pure self-play  \cite{silver2,silver2017mastering}
is  an \emph{effective integration of a learner and a solver.}
AlphaZero is  a modified, pointwise version of policy iteration, a basic planning algorithm for MDPs
and adversarial games where a policy is evaluated and improved iteratively  \cite{bertsekas:alphazero}.
In AlphaZero, a neural network is used to represent  the  value and 
policy functions, and in each iteration both  functions are
improved incrementally at one  state (and  at others by generalization) by carrying out a
Monte Carlo tree search \cite{coulom:mcts,uct}  from such state  guided by
the current value and policy functions.  Interestingly, AlphaZero is similar to an
algorithm developed independently  at the same time, Expert Iteration (ExIt), that is   explicitly cast as an integration of
System~1 and System~2 inference \cite{barber}. Both AlphaZero and ExIt can also be understood as  systems that
learn by  imitation  but with a planner as teacher,   and with the ability to iteratively improve the teacher.

Two key questions are what are  the inherent limitations of these  algorithms, and what else would
be needed in order to get a more \emph{general integration of System 1 and System 2 inference}
in AI systems.  As mentioned before, a key restriction of  learners relying on neural networks is that
the \emph{size  of their inputs $x$ is  fixed.} This implies  that learners cannot emulate solvers
even over specific domains.  That is, deep learners cannot emulate a classical planner
or a  domain-specific Blocks world planner unless arbitrary instances can be expressed in finite size.
Attention mechanisms have been proposed for this  but attention may be more
relevant for executing  policies than for  computing them.  Regarding the second question,
some key dimensions for a more \emph{general and synergistic integration of learners and solvers}
are:

\bigskip

\noindent \textbf{Model learning.} Solvers provide flexibility and generality, but solvers need models.
Explanation and accountability also require  models:  we build explanations and make predictions using  models
\cite{darwiche,josh,pearl}.  Model-based reinforcement learning is  model-learning  but the  standard algorithms  assume
that the state variables are given.   Learning models from streams of actions and partial observations remains challenging. 

\medskip

 \noindent  \textbf{Learning the relevant variables.} In deep learning, representation learning is learning functions (features) of the inputs that are somehow
 reusable. Features that are particularly reusable are the \emph{variables of a problem}. For example, many of the Atari games
 involve 2D objects that move, and things that happen as a result of  collisions. How to learn such  variables from the screen?
 Should  the moving objects  emerge from  a general learning scheme, from prior knowledge,  or  from a  combination of both
 \cite{kuipers:pong,bengio:features,marcus2}? 

 \medskip
 
 \noindent  \textbf{Learning finite-size  abstract  representations.} The input size restriction makes it challenging for DRL methods to learn
 general policies for achieving a goal like $on(A,B)$ over all instances with blocks $A$ and $B$. An alternative to the use of attention mechanisms
 is to learn  problem   abstractions of bounded  size  that depend  on the  goal. For example, for achieving the   goal  $on(x,y)$
 the only  features that matter  are  the number of blocks above $x$ and above $y$, along with some booleans, 
 and a convenient, fixed-size  abstraction of this  family of problems  can  be obtained by projecting the actions
 over such features \cite{bonet:features}.  The challenge is to learn such  features and projections automatically.

 \medskip

 Most work in DRL has not been aimed at an integration of learners and solvers but at expanding the range of  problems  handled by DRL.
 Something similar can be said about work in planners and solvers. Yet, the solutions  to some of the bottlenecks faced by each 
 approach, that often have to do with representational issues,  may lie in the integration that is  required  to resemble   the  integration
 of Systems~1 and 2 in the human mind. One  area where research on solvers and learners  meet, as mentioned above, is in the computation of
 generalized plans  capable of solving many  problems at the same time.  Other types of challenges are dealing with  other agents, virtual or human,
 in what  constitutes the social dimension of intelligence, and dealing with a physical and social  world that is not segmented
 into problems or tasks  \cite{dietterich:innateness}. 
  
\section{AI: Dreams and Nightmares}

AI is not close to a general, human level-intelligence  but this does not mean that current and future AI capabilities can't 
be used for good or ill. The Asilomar AI Principles enumerate  ways in which  harms can be avoided and benefits
realized\footnote{https://futureoflife.org/ai-principles.} but the principles are  not easy to enforce. 
It would be great  to align the values of  AI systems with human values,  but the same applies to technology in
general, politics, and economics. Clearly, there are other forces at play: forces like  a powerful market  and a
weak  political system, both increasingly aimed at our System~1.  I have argued why flexible and transparent
AI systems need a System 2  that  cannot be provided by learners alone. The same   applies to our collective lives
where learning from mistakes, when it is  possible at all, can be  very costly.
For living together in modern societies, we need a well-functioning
System~2 informed by  facts  and the common good.\footnote{\cite{stanovich:robot,heath:reason}.}
If we  want good AI, we can't look away from culture and politics.


\medskip

\section*{Acknowledgments}

I thank J\'{e}r\^{o}me Lang for the invitation to speak at IJCAI,
Guillem Franc\`{e}s and Blai Bonet for many useful comments, and my
current and former students for the opportunity to learn with them.
The work is partially funded by grant TIN2015-67959-P, MINECO, Spain.



\bibliographystyle{named}
\bibliography{control}

\end{document}